\documentclass[conference]{IEEEtran}
\IEEEoverridecommandlockouts
\usepackage{cite}
\usepackage{amsmath,amssymb,amsfonts}
\usepackage{algorithmic}
\usepackage{graphicx}
\usepackage{textcomp}
\usepackage{xcolor}
\usepackage{algorithm,algorithmic}
\usepackage{subcaption}
\usepackage{comment}
\def\BibTeX{{\rm B\kern-.05em{\sc i\kern-.025em b}\kern-.08em
    T\kern-.1667em\lower.7ex\hbox{E}\kern-.125emX}}

\usepackage{url}
\PassOptionsToPackage{hyphens}{url}

\begin{document}

\title{Fast and Resource-Efficient Object Tracking on Edge Devices: A Measurement Study}

\author{\IEEEauthorblockN{Sanjana Vijay Ganesh$^{+}$, Yanzhao Wu$^{*,+}$, Gaowen Liu$^{**}$, Ramana Kompella~$^{**}$, Ling Liu$^{+}$}
$^{+}$ Georgia Institute of Technology, Atlanta, GA, USA \\
$^{*}$ Florida International University, Miami, FL, USA \\
$^{**}$ Cisco Systems, Inc. 170 West Tasman Dr. San Jose, California, USA
}

\maketitle

\begin{abstract}
Object tracking is an important functionality of edge video analytic systems and services. Multi-object tracking (MOT) detects the moving objects and tracks their locations frame by frame as real scenes are being captured into a video. However, it is well known that real time object tracking on the edge poses critical technical challenges, especially with edge devices of heterogeneous computing resources. This paper examines the performance issues and edge-specific optimization opportunities for object tracking. We will show that even the well trained and optimized MOT model may still suffer from random frame dropping problems when edge devices have insufficient computation resources. We 
present several edge specific performance optimization strategies, collectively coined as EMO, to speed up the real time object tracking, ranging from window-based optimization to similarity based optimization. Extensive experiments on popular MOT benchmarks demonstrate that our EMO approach is competitive with respect to the representative methods for on-device object tracking techniques in terms of run-time performance and tracking accuracy. EMO is released on Github at \url{https://github.com/git-disl/EMO}.
\end{abstract}

\begin{IEEEkeywords}
\noindent Object Tracking, Multi-object Tracking, Adaptive Frame Skipping, Edge Video Analytics. 
\end{IEEEkeywords}

\section{Introduction}
Video cameras are widely deployed on cellphones, vehicles, and highways, and are soon to be available almost everywhere in the future world, including buildings, streets and various types of cyber-physical systems. We envision a future where edge sensors, such as cameras, coupled with edge AI services will be pervasive, serving as the cornerstone of smart wearables, smart homes, and smart cities.
However, most of the video analytics today are typically performed on the Cloud, which incurs overwhelming demand for network bandwidth, thus, shipping all the videos to the Cloud for video analytics is NOT scalable, not to mention the different types of privacy concerns. Hence, real time and resource-aware object tracking is an important functionality of edge video analytics. Unlike cloud servers, edge devices and edge servers have limited computation and communication resource elasticity. 
This paper presents a systematic study of the open research challenges in object tracking at the edge and the potential performance optimization opportunities for fast and resource efficient on-device object tracking. 

Object tracking is a representative computer vision task to detect and track objects throughout the video~\cite{ObjectTracking-Survey,FairMOT,b4,SORT}. Multi-object tracking is a subgroup of object tracking that tracks multiple objects belonging to one or more categories by identifying the trajectories as the objects move through consecutive video frames. Multi-object tracking has been widely applied to autonomous driving, surveillance with security cameras, and activity recognition. A popular paradigm for MOT is tracking-by-detection~\cite{b4, SORT}, which first detects objects by marking them with object class labels and bounding boxes, computes the similarity between object detections, and associate tracklets by assigning IDs to detections and tracklets belonging to the same object. 
Online object tracking aims to process incoming video frames in real time as they are captured. However, deep neural networks (DNNs) powered multi-object trackers are compute-intensive, e.g., using convolutional neural networks (CNNs), such as YOLOv3~\cite{Yolov3} and Faster RCNN~\cite{FasterRCNN}, for detecting objects. When deployed on edge devices with resource constraints, the video frame processing rate on the edge device may not keep pace with the incoming video frame rate. This mismatch can result in lag or dropped frames~\cite{ParallelDetection}, ultimately diminishing online object tracking quality.

In this paper, we focus on reducing the computational cost of multi-object tracking by selectively skipping detections while still delivering comparable object tracking quality.
{\it First,} we analyze the performance impacts of periodically skipping detections on frames at different rates on different types of videos in terms of accuracy of detection, localization, and association.
{\it Second,} we introduce a context-aware skipping approach that can dynamically decide where to skip the detections and accurately predict the next locations of tracked objects.
{\it Third,} we conduct a systematic experimental evaluation on the MOTChallenge datasets~\cite{MOT15,MOT16}, which demonstrates that the proposed approach can effectively reduce computation costs of multi-object tracking by skipping detections and maintain comparable multi-object tracking quality to the no skipping baseline.

\section{Related Work}
\subsection{Categories of General Tracking Techniques}
\noindent \textbf{Batch Methods:} Some of the early solutions to object tracking use batch methods for tracking the objects in a particular frame, the future frames are also used in addition to current and past frames. The association stage is then formulated and solved using methods like the min-cost flow algorithm, and shortest path algorithm~\cite{global-data-association-MOT-network-flows,globally-optimal-tracking-MOT,MOT-k-shortest-paths}.

\noindent \textbf{Separate Detection and Embedding:} Some approaches like SORT~\cite{SORT} utilize a deep convolutional neural network for object detection and rely entirely on the shape, size, and location of bounding boxes for association. A few studies extended these approaches by using another model trained separately to extract appearance features or embeddings of objects for association. For example, \cite{DeepSORT,POI} leverage a detection model to identify the bounding boxes containing objects and another re-identification (Re-ID) model to extract the features of each bounding box to associate the object detections with existing tracks. 

\noindent \textbf{One-shot Trackers:} \cite{JDE,FairMOT,TrackRCNN} use a single shot DNN in a multi-task learning setup to output the bounding boxes and the appearance embeddings of the detected bounding boxes simultaneously for tracking objects.

\noindent \textbf{Improvements in Association Stage:} Several studies enhance object tracking quality with improvements in the association stage. \cite{DeepSORT,MOTDT,ByteTrack} introduce different forms of cascaded matching strategies that can improve the object association and IDF1 score~\cite{IDmetrics}. 
\cite{Trackformer,MOTR} adopt an attention mechanism that uses queries to compute the location of the tracked object in the next frame thus causing implicit matching. 

\noindent \textbf{Other Tracking Paradigms:} \cite{Trackformer} formulates the problem as a Markov Decision Process and uses Reinforcement Learning (RL) to decide the appearance and disappearance of object tracklets.

\subsection{Existing Representative Approaches and Limitations}
\subsubsection{Tracking Approaches Aimed towards Higher Tracking Accuracy}
SORT~\cite{SORT} performs detection with Faster-RCNN, position estimation with Kalman Filter, and association with Hungarian algorithm using bounding box IoU as a measure. It does not use object appearance features for association. The approach is fast but suffers from high ID switches. DeepSORT~\cite{DeepSORT} extends SORT~\cite{SORT} by using a separate ResNet model for extracting appearance features for re-identification. The track age and Re-ID features are also used for association, leading to a significant reduction in the number of ID switches but at a slower processing rate.
Track-RCNN~\cite{TrackRCNN} extends Mask-RCNN~\cite{MaskRCNN} (for segmentation) by adding a Re-ID head on top of Mask R-CNN. JDE (Joint Detection \& Estimation)~\cite{JDE} is an extension of YOLOv3~\cite{Yolov3} (used for object detection). JDE uses a single shot DNN in a multi-task learning setup to output the bounding boxes and the appearance embeddings of the detected bounding boxes simultaneously thus reducing the amount of computation needed compared to DeepSORT. FairMOT\cite{FairMOT} also uses a single CNN model for detection and re-identification in a multi-task learning setup. However, it uses an anchor-free detector that predicts the object centers and sizes and extracts Re-ID features from object centers. 

Several studies focus on the association stage. ByteTrack~\cite{ByteTrack} adds a second step to the association stage. In addition to matching the bounding boxes with high scores, it also recovers the true objects from the low-scoring detections based on similarities with the predicted next position of the object tracklets. Observation-centric SORT~\cite{ObservationCentricSORT} aims to overcome the limitations of Kalman filter in scenarios where objects move non-linearly. It uses a smoothing approach based on observations to remove the accumulated error after it recovers from occlusion and shows good performance on the DanceTrack dataset~\cite{DanceTrack} in which objects exhibit non-linear and abrupt movements. BoT-SORT~\cite{BoTSORT} and Deep OC-SORT~\cite{DeepOCSORT} leverage camera motion compensation for association in addition to motion and appearance features. BoT-SORT introduces a more accurate Kalman filter state vector. Deep OC-SORT employs adaptive re-identification using a blended visual cost. MotionTrack~\cite{MotionTrack} learns robust short-term interaction-aware motions and long-term motions to recover from extreme occlusions using the history trajectory of the target object in a unified framework to associate trajectories from a short to long range.
TrackFormer~\cite{Trackformer} introduces a new paradigm by using an attention-based model to jointly perform detection and tracking. Tracktor~\cite{Tracktor} uses the regression head of the Faster RCNN object detector to align the position of an object on one frame to the new position of this object on the next frame. This approach does not require a separate tracker, where classification scores are used to decide whether to kill occluded tracks. GSDT~\cite{GSDT} performs joint detection and association using Graph Neural Networks (GNNs) to model the relations between objects in spatial and temporal domains.

\subsubsection{Approaches Focused on High Speed and Low Computational Cost}
\cite{speed} performed a detailed analysis by comparing the tracking performance of SORT for videos of different input frame rates. This study shows that it leads to a significant drop in accuracy by sampling the video at a low frame rate to deliver real-time tracking. HTracker~\cite{HTracker} employs a static skipping approach and skips detection on every alternate frame. HTracker uses a CNN-based detector to perform object detection for non-skipped frames and apply the particle filter estimate of the coordinates for the skipped ones. This approach is sub-optimal, especially for videos where objects do not exhibit uniform motion. For some segments with slow object motions, skipping more frames may have minimal impact on accuracy while in some other segments with fast motions, skipping even a single frame may result in a significant loss in accuracy. 
Detect-or-Track~\cite{DetectOrTrack} uses an adaptive skipping approach where a Siamese network is used to predict similarity between consecutive frames to determine whether to skip detection of the specific frame. When detection is skipped, a deep feature extractor is used to identify the nearest patch with the highest similarity to the object in the bounding box as the tracking result. \cite{ECRT} attempts to reduce the load on edge devices by intelligently partitioning CNN inference into two parts, which are executed locally on an IoT device and/or on the edge server.

\subsubsection{Approaches that Use Frame Skipping for Other Video Processing Applications}
Similar frame-skipping approaches have been applied to other video processing applications. FrameHopper~\cite{FrameHopper} introduces an approach to select frames to be sent to the cloud for detection-driven video analytics. It uses an RL agent on the edge device to estimate how many frames can be skipped. \cite{IFSM} uses an intelligent frame skipping mechanism for video streaming and reconstruction, which leverages an estimate of the motion between frames to decide whether the frame can be skipped or not.

\begin{table}[htbp]
\caption{Existing representative approaches}
\begin{center}
\begin{tabular}{|p{3.5cm}|p{4.5cm}|}
\hline
\textbf{Approach}&\textbf{Limitations} \\
\hline
Batch approaches (e.g.,~\cite{global-data-association-MOT-network-flows}) &
Predictions are not done in real-time. The algorithm will be able to output the predictions for frames only after several future frames are obtained. \\
\hline
Approaches the use separate models for detection and Re-ID features &
Difficult to reach real-time performance due to the use of more than one model \\
\hline
Approaches that use public detections (e.g.,~\cite{Neural-Solver-MOT})
 &
Do not take the detection time into account. They only perform the association step. In practical applications, the performance of the entire system (detection $+$ association) is important.
\\
\hline
Approaches that perform association using spatial overlap alone (e.g.,~\cite{SORT} \& \cite{IOUTracker}) &
Do not work well in cluttered scenes and fast camera motion due to not using Re-ID features. \\
\hline
Static Skipping (e.g.,~\cite{HTracker}) &
Always skip a specific number of frames each time. It is possible to skip more depending on the video context. \\
\hline
Other approaches for dynamic skipping (e.g.,~\cite{DetectOrTrack}) &
The use of DNNs to compute similarity is compute-intensive. \\
\hline
\end{tabular}
\label{tab3}
\end{center}
\end{table}

\section{Object Tracking Overview}

Multi-object tracking (MOT) aims to detect multiple objects and track their trajectories (or stationary positions) in a video.
This work aims at reducing the computational cost of MOT models so that it can run in real-time on edge devices with limited computing capacity. We explore how to leverage frame skipping techniques to reduce the computation costs of multi-object tracking without compromising tracking accuracy.
The input to MOT is a sequence of raw video frames from an RGB camera captured at a specific frame rate (frames per second (FPS)). The input size for each video will be the Number of frames $\times$ Width of the image $\times$ Height of the image $\times$ Number of color channels ($=$3). The output is the list of objects belonging to the categories under consideration and their trajectories across frames. For each frame, the format of the output is a list of 2D bounding box coordinates (left, right, height, width) of each object detected and the ID of the object. This ID uniquely identifies the object across different frames.
In real-world object tracking scenarios, ground truth object detections are not available. Therefore, in this study, we do not use the provided public detections by the MOTChallenge datasets~\cite{MOT16}.
Also, this work aims at achieving real-time object tracking and hence focuses only on online MOT approaches where information from current and past will be used while processing a particular frame, excluding future frames.

Tracking-by-detection is a popular paradigm that is widely used by many object tracking approaches~\cite{b4,SORT}.
This paradigm primarily consists of three stages, (1) detection, (2) prediction, and (3) association.
The detection stage identifies the objects of interest in each frame and localizes them using an object detector, such as a single stage object detector like RetinaNet~\cite{RetinaNet}, CenterNet~\cite{CenterNet}, YOLO~\cite{Yolov3} or a two-stage object detector like Faster R-CNN~\cite{FasterRCNN}. The prediction stage predicts the next locations of the object tracks. Common approaches for this stage include optical flow~\cite{b8}, recurrent neural networks (RNNs)~\cite{b9,b10}, Kalman filter~\cite{SORT,DeepSORT}, and particle filter~\cite{b12,b13}. RNN can track the motion and interactions of target objects for a longer period of time, making it suitable for the presence of long-term occlusions~\cite{b14}. Kalman filter~\cite{kalman} is used to estimate the next state of a linear dynamic system, which models the velocity of the objects to compute an estimation of the next position of the object by combining the previous estimate with new observations. The particle filter performs state estimation of a non-linear dynamic system by using a set of particles (samples) to approximate the probability distribution of the next state.
The final stage is to associate the detected objects across frames. This stage uses the similarity between predicted locations of existing tracks and detections in the next frame to associate the object detections to existing object tracks. MOT approaches use Hungarian algorithm~\cite{hungarian} or Bipartite graph matching for optimal assignment problem (one-to-one matching with minimum cost) to compute the optimal matching between the tracked objects and the detections in the next frame. Intersection-over-Union (IoU), and appearance features are used to compute the cost matrix that defines the cost between each detection and prediction. Trackers based on Intersection over Union (IoU)~\cite{IOUTracker, SORT} utilize the distance and similarity between detected and predicted bounding boxes in terms of location, size, and shape. In order to compute appearance features from the portion of the image containing the object of interest, several approaches such as color histograms, Histogram of Oriented Gradients(HOG), learning object motion~\cite{Transtrack} and re-identification features extracted using CNNs~\cite{DeepSORT, FairMOT, JDE, CSTrack} have been explored. Some approaches also use a cascaded matching strategy that matches the most recent tracks then lost ones~\cite{DeepSORT}: matches based on appearance similarity and then matches using IoU~\cite{MOTDT}. Some other approaches~\cite{Trackformer} use an attention mechanism that provides implicit matching through the use of queries to compute the location of the tracked object in the next frame.

\begin{figure*}[ht]
  \centering  
  \vspace{-5ex}
  \includegraphics[width=\textwidth, height=0.80\textheight]{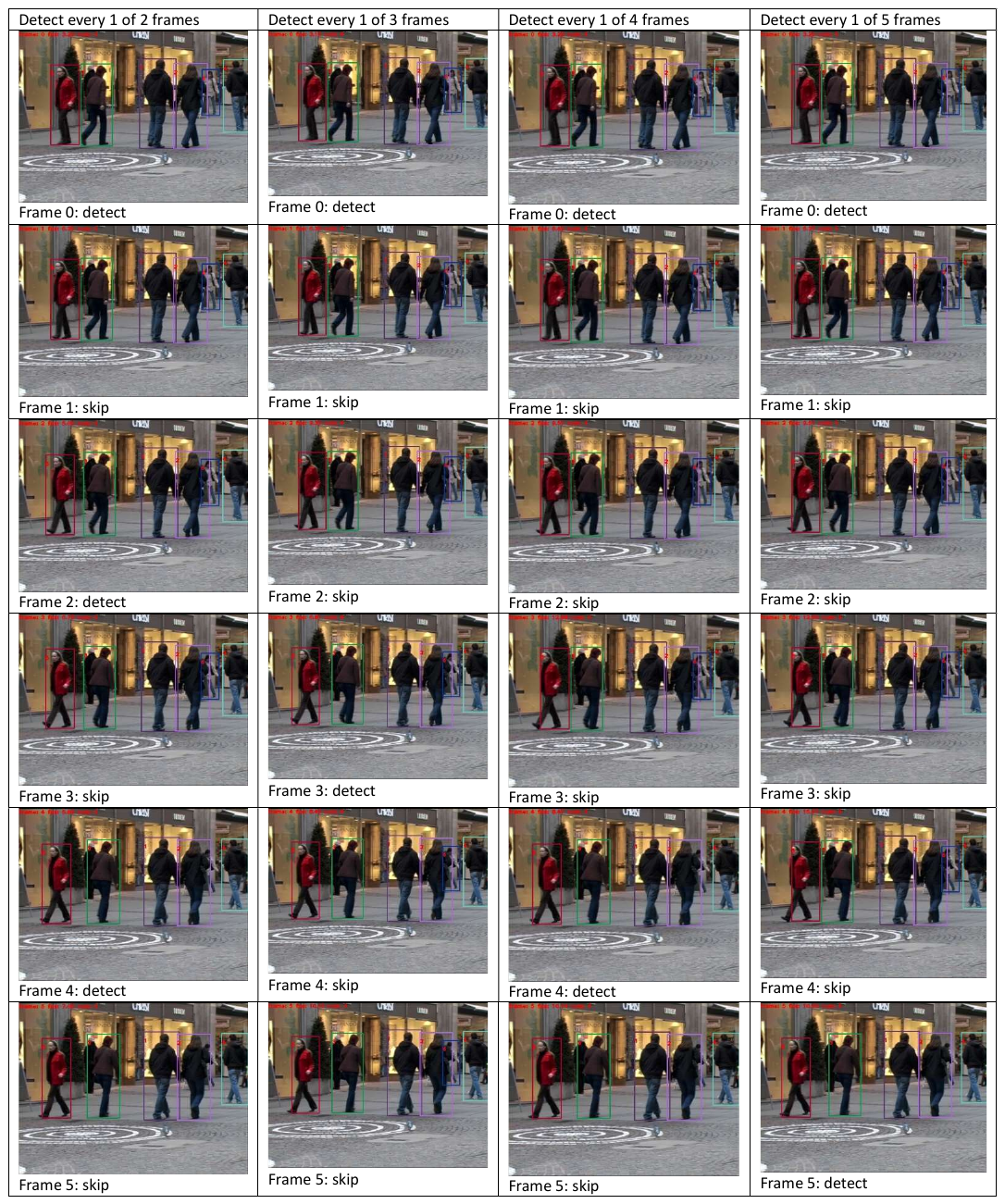}
  \vspace{-27ex}
  \caption{Each column shows the tracking results of six consecutive frames from TUD-Stadtmitte (25 FPS) from MOT-15 dataset when the detections are skipped at different frequencies and the bounding boxes from the previous detections are reused. It is observed that for this subsequence, upto 3 frames can be skipped without much degradation in the quality of object tracking.
  }
  \label{fig:TUDS_5_sample_frames_periodic}
\end{figure*}

\subsubsection{Optimization Strategies}
In the tracking-by-detection paradigm, detection is performed on each frame. Since the accuracy of the detector is critical for MOT approaches adopting tracking-by-detection, most of the recent approaches use a CNN-based object detector which is computationally intensive. Considering a 25 FPS video to be processed in real-time, the device will have to run inference on a CNN for at least 25 frames each second. On a device with low computational capability, it may not be feasible to achieve this real-time performance~\cite{ParallelDetection}.
On the other hand, in many real-world applications like surveillance, it could be possible to skip detections on some frames and still be able to reach comparable accuracy. In some circumstances, the objects do not move from one frame to another, for example, feed from a surveillance camera in a parking garage. In such cases, the objects in the view of the camera remain stationary for a long time and the detection results will remain the same for several consecutive frames. This makes it possible to skip detection on frames in between and save on computations without loss in quality of object tracking.

\begin{figure*}[ht]
  \centering  
  \vspace{-5ex}
  \includegraphics[width=\textwidth, height=0.80 \textheight]{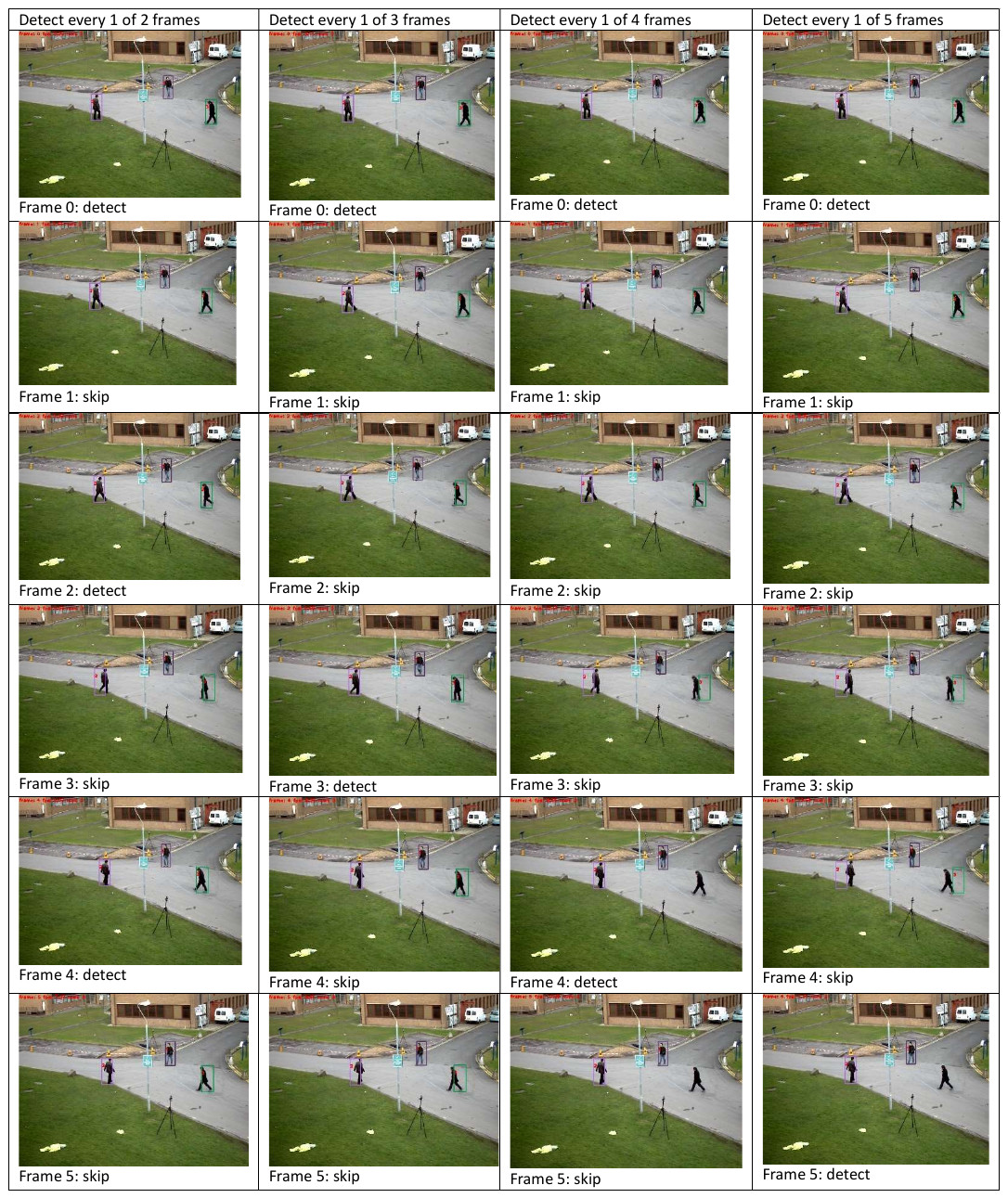}
  \vspace{-27ex}
  \caption{Five consecutive frames from PETS09-S2L1 from MOT-15 dataset captured at 7 FPS and the tracking results when the detections are skipped at different frequencies and the bounding boxes from the previous detections are reused.
  }
  \label{fig:PETS_5_sample_frames_periodic}
\end{figure*}

When the objects detected move slowly and the video is captured at a high frame rate of 25 or 30 FPS, the motion of the object or the change in orientation of the object between successive frames will be very small, as observed in the image frames in Figure~\ref{fig:TUDS_5_sample_frames_periodic}. In these frames, in the last column where 4 consecutive frames are skipped and the detected bounding boxes for frame 0 are reused for all the remaining 5 frames, it can be observed that only in the fifth frame do the people start to get outside the box, making it reasonable to reuse the same bounding boxes for at least 3 frames. Since successive frames would look very similar, this allows skipping detection on multiple frames without much loss in tracking accuracy.

When the objects detected move fast in such a way that the movement between consecutive frames is significant, skipping detections on these frames may result in a drastic decrease in the object tracking quality. Skipping detections on multiple frames might cause the tracking algorithm to miss a new object entering the scene or a stationary object starting to move. Moreover, when the frame rate of the video is small (around 10 FPS), the movement between consecutive frames is often much larger. In the frames in the last two columns of Figure~\ref{fig:PETS_5_sample_frames_periodic}, 3 or 4 consecutive frames are skipped, and the detected bounding boxes for frame 0 are drawn for all 4 or 5 frames. We observe that within three frames, the people start to get outside the bounding box, making it reasonable to reuse the same bounding box for up to 2 frames (as can be seen in the first two columns of Figure~\ref{fig:PETS_5_sample_frames_periodic}). In these circumstances, within the short duration of a small number of frames, an object could enter or leave the scene or get occluded to a considerable extent. In these circumstances, skipping detection even for a small number of consecutive frames will cause a considerable degradation in the quality of tracking. When the video is captured from a moving camera, such as from a drone, the moving objects and the stationary objects in the background exhibit greater movement compared to that captured from a stationary camera. 
If we assume that the detections on each frame are perfectly accurate, within the frames on which detections were skipped, an object might get occluded and/or might come out of an occlusion or a detected object might change its orientation significantly. This could lead to changes in re-identification features or the position of a particular track relative to a nearby track which could potentially lead to the object being associated with another nearby track. This increases the number of ID switches and reduces the IDF1 score.

\subsubsection{Categories of Techniques}
The frame rate of the video, speed of motion of the objects of interest, and whether the video is taken from a static/dynamic camera are important factors to determine the frames on which detections can be skipped with minimal loss in the object tracking quality.
A baseline method to implement the detection skipping would be to follow a periodic frame skipping approach like~\cite{HTracker} and perform detections once in every $\omega$ number of frames, where $\omega$ can be empirically identified or heuristically calculated based on the factors mentioned above. However, the periodic skipping approach is suitable only when the speed of motion of the objects remains nearly the same throughout the video. In the surveillance application, there might be some segments in the video that remain majorly static while other segments in the same video capture fast moving objects. In these scenarios, a single frame skipping rate may not work well for the entire video while an aperiodic skipping scheme such as~\cite{DetectOrTrack, FrameHopper} may achieve better performance.
The semantics of objects between successive frames can serve as an indicator to determine which frame to skip, such as the similarity between two frames. For example, if the next frame is very similar to the previously detected frame, the next frame can be skipped.
Traditional image processing techniques can be used for extracting semantics as well as alternative approaches, including using a neural network~\cite{DetectOrTrack} or using reinforcement learning~\cite{FrameHopper}.

\subsubsection{Metrics Used for Efficiency Measurement}
Multi-object tracking involves detection, localization, and association. The classical metrics~\cite{ClassicalMetrics} to evaluate multi-object tracking approaches compute the number of trajectories that are correctly tracked in 80\% of the frames (Mostly Tracked - MT), number of trajectories that are correctly tracked in less than 20\% of the frames (Mostly Lost - ML), trajectories that are covered by more than one fragment, false trajectories and number of times when the ID associated with a correctly tracked object is wrongly changed (ID switches). The CLEAR MOT metrics~\cite{CLEARmetrics} were developed for the Classification of Events, Activities, and Relationships (CLEAR) workshops. These metrics are summarized in Table~\ref{tab:ClearIDMetrics}. MOTA and MOTP match the ground truth detections and tracks frame-by-frame. ID scores~\cite{IDmetrics} (ID Precision, ID Recall, ID F1) reward the tracker that follows the objects for the longest time possible by performing the mapping globally rather than on a frame-by-frame basis.

Since different CLEAR and ID metrics focus on different aspects of tracking (e.g., MOTA emphasizes detections and ID F1 emphasizes association), HOTA (Higher Order Tracking Accuracy)~\cite{HOTAmetrics} can be used as a single unified metric that combines the accuracy of detection, localization, and association in a balanced view. HOTA allows direct comparison and ranking of trackers and its sub-metrics (Detection Accuracy, Localization Accuracy, and Association accuracy) allow performance analysis of different components of the tracker. A brief summary of the HOTA and its sub-metrics are provided in Table~\ref{tab:HOTAMetrics}.

\begin{table*}[htbp]
\caption{\label{tab:ClearIDMetrics} Metrics used for efficiency measurement – CLEAR \& ID }
\begin{center}

\begin{tabular}{|p{2cm}|p{6cm}|p{3cm}|p{5cm}|}
\hline
\textbf{Metric}&\textbf{Formula}&\textbf{Purpose}&\textbf{Explanation} \\ \hline
MOTA ($\uparrow$) \newline Multi-object tracking accuracy &
 $MOTA = 1 - \frac{(FN+FP+IDSW)}{GT} \in (-\infty,1] $
 \newline
 FP-false positive; 
 \newline
 FN-False negative; 
 \newline
 IDSW-ID switches; 
 \newline
 GT – number of ground truth detections
 & 
Evaluates detections (FP \& FN) and association (IDs) (to some extent) but not localization &
Performs one-to-one matching at detection level between predicted and ground truth detections to compute FP, FN, IDSW. \\
\hline
MOTP ($\uparrow$) \newline Multi-object tracking precision &
 $\frac{ \text{Sum of bounding box  overlap} \text { between predictions and ground truths} }{\text{number of matches (True Positives)}}$ & 
Evaluates localization performance &
MOTP averages the overlap between all correctly matched predictions and their ground truth. \\
\hline
ID F1 ($\uparrow$) \newline Identification F1 &
Identification Precision $IDP = \frac{IDTP}{IDTP + IDFP} $
 Identification recall $IDR = \frac{IDTP}{IDTP + IDFN} $
 Identification F1 $IDF1 = \frac{2} {\frac{1}{IDP} + \frac{1}{IDR}} = \frac{2 IDTP}{2 IDTP + IDFP + IDFN} $
 & 
Evaluates association accuracy &
Uses Hungarian algorithm to compute one-to-one mapping between predicted and ground truth trajectories. \\
\hline
IDs ($\downarrow$) \newline ID Switches &
 & 
Evaluates association&
The number of times the ID of tracks are swapped (after occlusion or when the pass close to each other). \\
\hline
\end{tabular}
\label{tab1}
\end{center}
\end{table*}

\begin{table*}[htbp]
\caption{\label{tab:HOTAMetrics} Metrics used for efficiency measurement - HOTA }
\begin{center}
\begin{tabular}{|p{1.5cm}|p{5cm}|p{3cm}|p{6.5cm}|}
\hline
\textbf{Metric}&\textbf{Formula}&\textbf{Purpose}&\textbf{Explanation} \\
\hline
LocA ($\uparrow$) \newline Localization Accuracy &
 $LocA = \frac{1}{|TP|} \sum_{c \in TP} Loc\text{-}IoU(c) $ & 
 Evaluates localization performance &
Averages the Loc-IoU over all pairs of matching predicted and ground-truth detections in the whole dataset. \\
\hline
DetA ($\uparrow$) \newline Detection Accuracy &
 $DetA = Det\text{-}IoU = \frac{|TP|}{|TP| + |FN| + |FP|}$
  & 
Evaluates detection performance &
Calculates Det-IoU using the count of TPs, FNs and FPs over the whole dataset. Uses a localization threshold to determine which detections overlap. Uses Hungarian algorithm for 1-1 matching. \\
\hline
AssA ($\uparrow$) \newline Association accuracy &
 $AssA = \frac{1}{|TP|} \sum_{c \in TP} Ass\text{-}IoU(c) = \frac{1}{|TP|} \sum_{c \in TP} \frac{TPA(c)}{TPA(c) + FNA(c) + FPA(c)} $
 & 
Measures how well a tracker links detections over time into the same identities &
The intersection between two tracks can be measured as the number of True Positive matches between the two tracks (Hungarian algorithm is used). AssA Averages the Ass-IoU over all pairs of matching predicted and ground-truth detections in the whole dataset. \\
\hline
HOTA ($\uparrow$) \newline High Order Tracking Accuracy &
$HOTA_{\alpha} = \sqrt{DetA_{\alpha} AssA_{\alpha}} = \sqrt{\frac{\frac{1}{|TP|} \sum_{c \in TP} Ass\text{-}IoU_{\alpha}(c)}{|TP_{\alpha}| + |FN_{\alpha}| + |FP_{\alpha}|}} $
$HOTA = \int_{0 < \alpha \leq} HOTA_{\alpha} \approx \frac{1}{19} \sum_{\alpha = 0.05; \alpha += 0.05} ^{0.95} HOTA_{\alpha}$
 & 
Unified metric that evaluates detection, localization and association & 
Combines a detection score and an association score by performing matches at the detection level while scoring association globally over trajectories. 
Final score is the geometric mean of the detection score and the association score. Then by integrating over the different $\alpha$ thresholds, we include the localization accuracy into the final score
 \\
\hline
\end{tabular}
\label{tab2}
\end{center}
\end{table*}

\section{Object Tracking with EMO}
\subsection{Motion Aware Periodic Skipping}
A baseline approach to reducing the computational cost of object tracking is to skip detections at random frames or to skip detections periodically. This approach avoids complex computations to decide which frames to skip. The periodic skipping approach will define a skipping window size $\omega$ and skip $\omega$ frames each time, i.e., running detections on 1 frame for every $\omega$ frames. The $\omega$ can be set as 1, 2, 3, 4, 5. 
However, the object tracking quality is sensitive to the hyperparameter $\omega$ (frequency of skipping), which requires careful tuning and depends on a number of factors, including the frame rate of video streams, object motions, and whether the video was captured from a fixed or moving camera. 
For example, if the objects in the video move fast (as shown in Figure~\ref{fig:PETS_5_sample_frames_periodic}) or in a non-linear direction, a larger $\omega$ value and hence a larger number of skipped frames will lead to a drop in object tracking accuracy. However, if $\omega$ is too small, though it might be suitable for videos that have fast or non-linear motion, it will incur high computation costs for processing videos where objects have linear and slow motion (as shown in Figure~\ref{fig:TUDS_5_sample_frames_periodic}), where a high number of frames could be dropped without impairing object tracking accuracy. 

\begin{table}[htbp]
\caption{Comparison of impact of skipping with different frequencies on the tracking performance on MOT-15 dataset}
\begin{center}

\begin{tabular}{|c|c|c|c|c|}
\hline
\textbf{$\omega$} & \textbf{MOTA ($\uparrow$)} & \textbf{MOTP ($\uparrow$)} & \textbf{IDF1 ($\uparrow$)} & \textbf{IDSW ($\downarrow$)} \\
\hline
No skip &	67.62\% &	0.206 &	75.09\% &	142 \\
\hline
1/3 frames skipped &	63.5 \% &	0.213	& 70.2 \% &	256 \\
\hline
1/2 frames skipped &	60.93\% &	0.229	& 68.46\% &	260 \\
\hline
2/3 frames skipped &	52.31\% &	0.247 &	60.74\% &	415 \\
\hline
\end{tabular}
\label{diffomega}
\end{center}
\vspace{-5mm}
\end{table}

\begin{figure}[htbp]
\centerline{\includegraphics[width=0.5\textwidth]{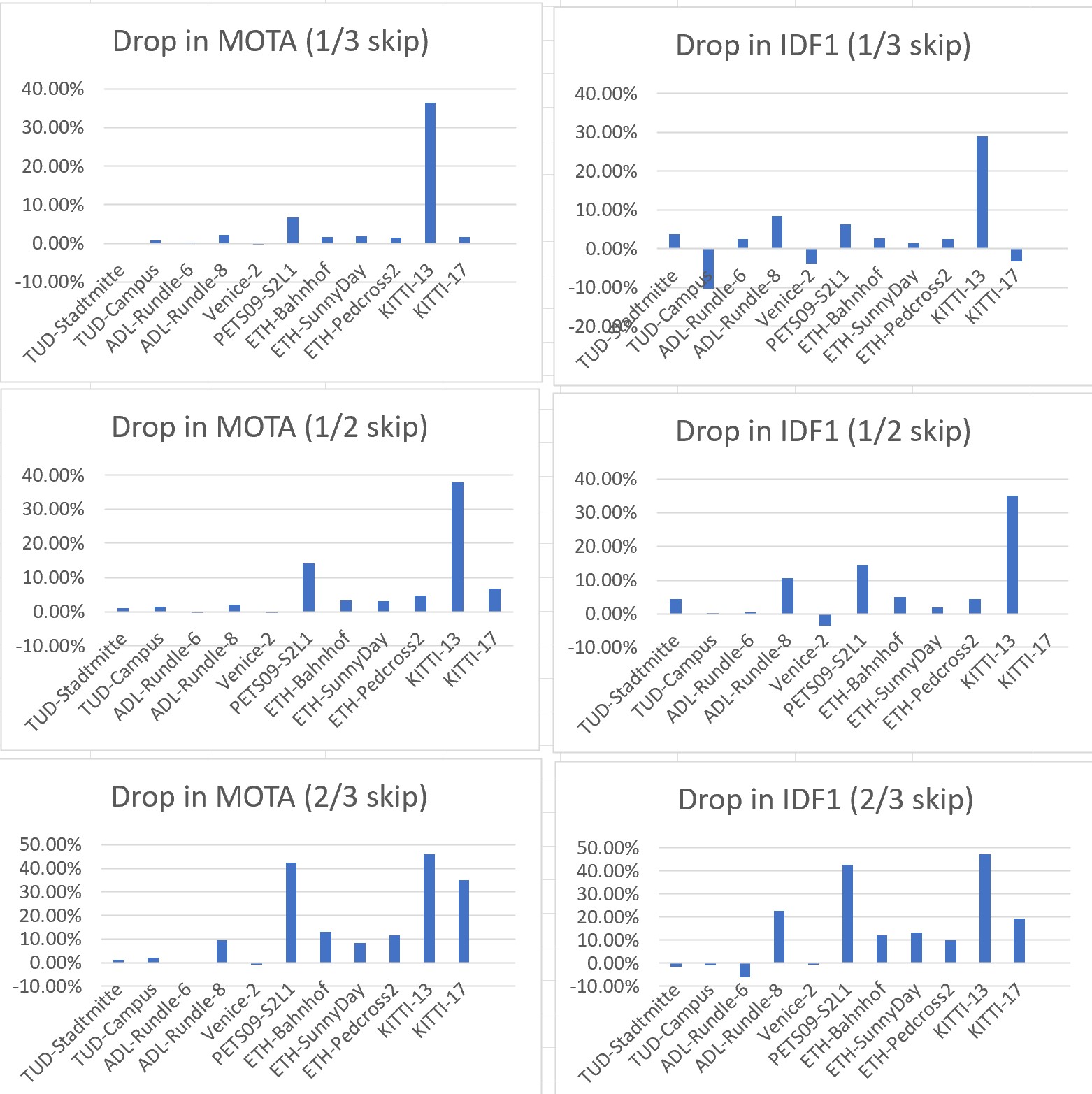}}
\caption{Plots showing the drop in MOTA, IDF1 while skipping detections on frames at different frequencies when compared the metrics calculated without skipping frames in different videos from MOT-15 dataset}
\label{drop_with_periodic}
\end{figure}

Table~\ref{diffomega} shows the performance of the FairMOT tracker on the MOT-15 dataset~\cite{MOT15} when frames are skipped at different frequencies and the bounding boxes from the latest detected frames will be reused for these skipped frames. Figure~\ref{drop_with_periodic} further shows the drop in MOTA, MOTP and IDF1 of each video due to frame skipping. The first 5 videos (first 5 bars along the x-axis) have a high incoming video frame rate of $>$ 20 FPS while the last 6 videos have a relatively low frame rate of $<$ 20 FPS. We observe that the videos with a lower FPS show a significant drop in object tracking accuracy (MOTA and IDF1) compared to the videos with a higher incoming FPS. Moreover, for the two videos KITTI-13 and KITTI-17 of the same video frame rate at 10 FPS, KITTI-17, which is taken from a static camera, shows a significant drop in accuracy only when more than half the frames are skipped, whereas KITTI-13, which is shot from a moving camera, shows a significant drop in accuracy even when 1/3rd of the frames are skipped.
Therefore, a constant $\omega$ may not work efficiently for various types of videos with different frame rates, such as shown in Figure~\ref{fig:TUDS_5_sample_frames_periodic} and~\ref{fig:PETS_5_sample_frames_periodic}.
Even within the same video, it would be beneficial to have different values of $\omega$, when the motion of objects is not always uniform and some segments allow more skipping than others.
Motion-aware periodic skipping addresses this problem by leveraging heuristics to determine the value of $\omega$ based on the FPS of the video and whether the video is taken from a static or a moving camera. However, this is still a sub-optimal solution, as it is challenging to come up with a single FPS threshold to provide optimal performance for all videos in practice. 
We introduce context-aware skipping to address this challenge.

\subsection{Context Aware Skipping}

Context-aware skipping approach aims to compute the number of frames to skip adaptively based on the video context. It performs object detections only on key frames that are considerably different from the previous frame.

\noindent \textbf{Identify Key Frames.} In order to identify these key frames, we compute the similarity of each incoming frame with the frame on which the last detection was performed. If the similarity is higher than a pre-defined threshold, detection on this new frame will be skipped and the bounding boxes from the last detections will be reused. Otherwise, the detection will be performed on this new frame.
Similarity can be computed on the full image or only over the areas covered by the bounding boxes.

\noindent \textbf{Predict Next Positions.} The Kalman filter can predict the movement of each tracked object, where the predicted next position of its bounding box can be used for skipped frames.
Most representative object tracking approaches use Kalman filters to predict the next position of each object tracklet and use that for associating the detected bounding boxes across consecutive frames.
If the Kalman filter estimate of the next positions of the tracklets on a particular frame is highly accurate, the detections on the next frame can be skipped. Considering the example of Figure~\ref{fig:PETS_5_sample_frames_periodic}, though the object moves significantly within 3 frames, the movement is still linear and can be accurately predicted by the Kalman filter. In such cases, more frames can be skipped with less loss of accuracy if we use the Kalman filter estimate for the frames where the detections are skipped. 

\begin{figure}[htbp]
\centerline{\includegraphics[width=0.5\textwidth]{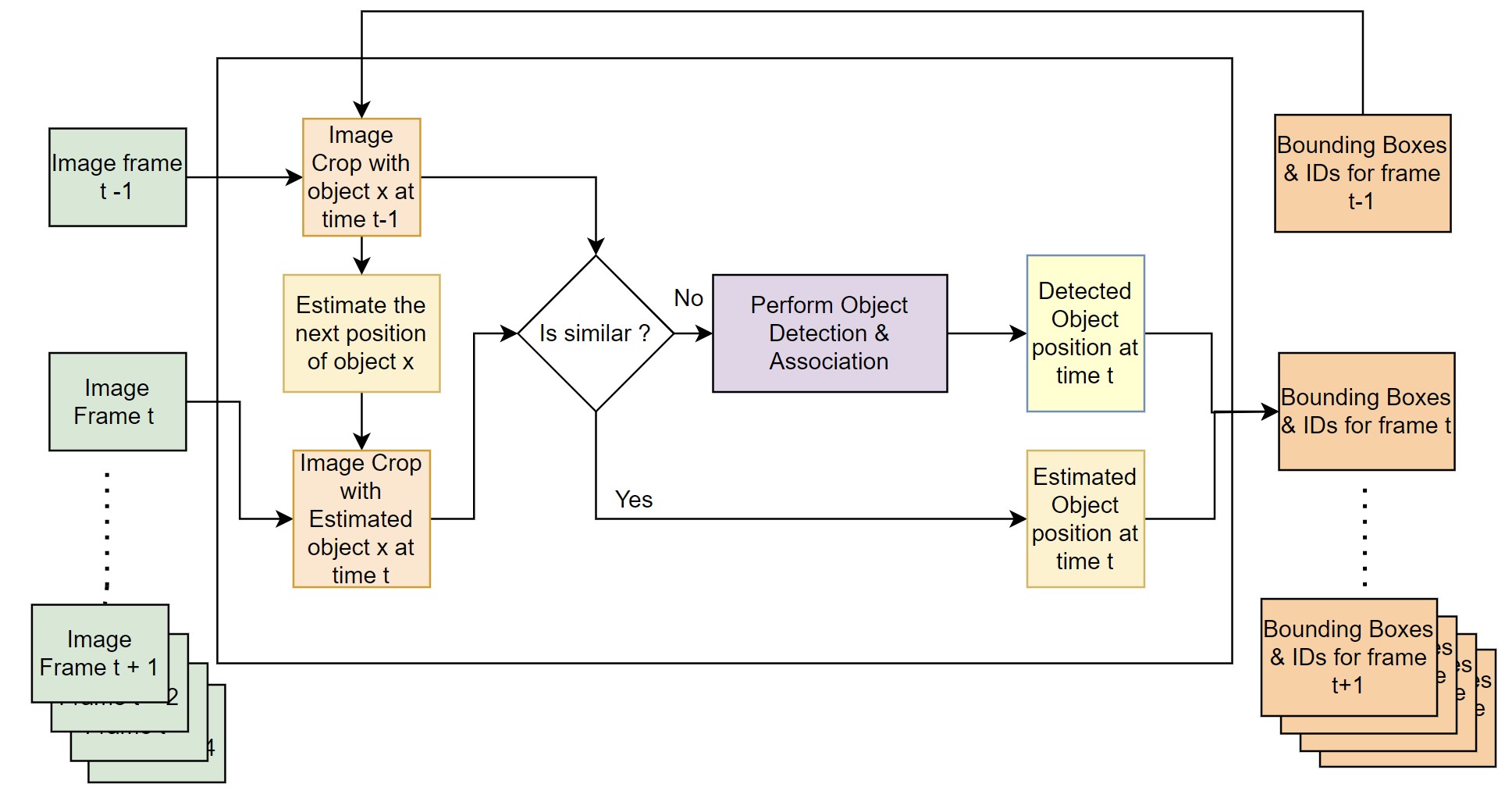}}
\caption{Block diagram of context aware skipping approach}
\label{block_diagram_EMO}
\end{figure}

Integrating the similarity-based key frame identification and Kalman filter based next position prediction will allow detections to be skipped on more frames without compromising object tracking accuracy. However, for an extreme case, if detections are skipped for too many consecutive frames when new objects enter the frame, the object tracker may not be able to capture them and even initialize new object tracks. 
Therefore, it becomes necessary to force the detections to be performed at least once for every $k$ frames. The value of $k$ can be determined based on the input video features. For example, we found that a higher $k$ for videos with higher frame rates and a lower $k$ for videos with lower frame rates can deliver good performance.
Figure~\ref{block_diagram_EMO} presents the workflow of the EMO approach. We below describe the two core functions: image similarity computation and state estimation.

\subsubsection{Image Similarity}

An intuitive method to compute image similarity is to leverage a Deep Convolution Neutral Network (DCNN) based feature extractor, trained to extract features of objects and identify if the two given image crops (a bounding box from the previous frame and an estimated bounding box in the next frame) are of the same object. However, this approach is compute-intensive. In the worst case, where the estimate is incorrect, and the detection can be skipped on very few frames, the feature extractor will add to the overall computation cost similar to a normal object tracker, decreasing the speed-up obtained by skipping frames. Therefore, we primarily explore low-cost methods for image similarity computation. 

\noindent \textbf{Eigenvalue based Similarity.}
\cite{eigensimilarity} demonstrates an approach to compute whether the given two images are similar or not. The approach first uses the gray level values of the two images at each pixel location to form gray-level pairs which are then used to form a correspondence map. The shape of this 2D gray level correspondence map will be a diagonal straight line for two identical images and non-linear for dissimilar images. The smaller eigenvalue of the covariance matrix of the pairs is used as a similarity measure. It will be zero for identical images and a large value for dissimilar images.
However, with this approach, it is difficult to compute a measure of how dissimilar the images are. For our case, a small amount of dissimilarity can be tolerated. As it can be seen from Tables~\ref{petsdiffeigenval}, \ref{tudsdiffeigenval}, and~\ref{tudcdiffeigenval}, different values of the eigenvalue thresholds (below threshold the frame is skipped) deliver good performance for different videos. We observe that this similarity measure gives different thresholds for tolerable dissimilarities for different videos, making it very difficult to come up with a consistent threshold across videos to decide whether to skip the frame or not.

\begin{table}[htbp]
\caption{Comparison of impact of skipping with eigenvalue based similarity measure on the tracking performance on PETS09-S2L1 video from MOT-15 dataset}
\begin{center}

\begin{tabular}{|p{3cm}|p{0.6cm}|p{0.6cm}|p{0.6cm}|p{0.5cm}|p{1cm}|}
\hline
\textbf{Approach} & \textbf{MOTA ($\uparrow$)} & \textbf{MOTP ($\uparrow$)} & \textbf{IDF1 ($\uparrow$)} & \textbf{IDSW ($\downarrow$)} & \textbf{Frames skipped } \\
\hline
No skip &	89.8\% &	0.250 &	87.4\% &	9 & 0/795 \\
\hline
Skip based on Eigenvalue similarity over whole image (Threshold 60) &	84.3\% &	0.263 &	80.6\% &	14 & 203/795 \\
\hline
Skip based on Eigenvalue similarity over whole image (Threshold 60) with detections mandated once every 4 frames &	84.4\% &	0.263 &	79.7\% &	15 & 201/795 \\
\hline
Skipping based on similarity with crop predicted by kalman filter (threshold for total 3500) &	88.6\% &	0.252 &	90.4\% &	8 & 86/795 \\
\hline
Skipping based on similarity with crop predicted by kalman filter (threshold for total 4000) &	88.9\% &	0.252 &	87.5\% &	7 & 115/795 \\
\hline
\end{tabular}
\label{petsdiffeigenval}
\end{center}
\end{table}

\begin{table}[htbp]
\caption{Comparison of impact of skipping with eigenvalue based similarity measure on the tracking performance on TUD-Stadtmitte video from MOT-15 dataset}
\begin{center}

\begin{tabular}{|p{3cm}|p{0.6cm}|p{0.6cm}|p{0.6cm}|p{0.5cm}|p{1cm}|}
\hline
\textbf{Approach} & \textbf{MOTA ($\uparrow$)} & \textbf{MOTP ($\uparrow$)} & \textbf{IDF1 ($\uparrow$)} & \textbf{IDSW ($\downarrow$)} & \textbf{Frames skipped } \\
\hline
No skip &	78.4\% &	0.248 &	81.6\% &	4 & 0/179 \\
\hline
Skip based on Eigenvalue similarity over whole image (Threshold 100) &	76.2\% &	0.255 &	80.1\% &	2 & 115/179 \\
\hline
Skip based on Eigenvalue similarity over whole image (Threshold 200) with detections mandated once every 4 frames &	75.8\% &	0.265 &	83.9\% &	1 & 68/179 \\
\hline
Skipping based on similarity with crop predicted by kalman filter (threshold for total 1400) &	77.9\% &	0.247 &	77.8\% &	2 & 100/179 \\
\hline
\end{tabular}
\label{tudsdiffeigenval}
\end{center}
\end{table}

\begin{table}[htbp]
\caption{Comparison of impact of skipping with eigenvalue based similarity measure on the tracking performance on TUD-Campus video from MOT-15 dataset}
\begin{center}

\begin{tabular}{|p{3cm}|p{0.6cm}|p{0.6cm}|p{0.6cm}|p{0.5cm}|p{1cm}|}
\hline
\textbf{Approach} & \textbf{MOTA ($\uparrow$)} & \textbf{MOTP ($\uparrow$)} & \textbf{IDF1 ($\uparrow$)} & \textbf{IDSW ($\downarrow$)} & \textbf{Frames skipped } \\
\hline
No skip &	76.6\% &	0.198 &	71.9\% &	3 & 0/71 \\
\hline
Skip based on Eigenvalue similarity over whole image (Threshold 200) &	76.0\% &	0.212 &	71.0\% &	4 & 32/71 \\
\hline
Skipping based on similarity with crop predicted by kalman filter (threshold for total 1700) &	75.8\% &	0.199 &	75.5\% &	5 & 20/71 \\
\hline
\end{tabular}
\label{tudcdiffeigenval}
\end{center}
\end{table}

\noindent \textbf{Normalized Cross Correlation (NCC) based similarity.}
The normalized cross correlation method of template matching can be used for computing image similarity~\cite{NCC}. Given a template $t$ of size $N_x \times N_y$, the normalized cross correlation value or Pearson correlation co-efficient for an image $f$ at a position $(u, v)$ is computed using the following formula:
\begin{equation}
r=\frac{\sum_{x,y}{(f(x,y)-\bar{f}_{u,v})(t(x-u, y-v)-\bar{t})}}{\sqrt{\sum_{x,y}{(f(x, y)-\bar{f}_{u,v})^2}\sum_{x,y}{(t(x-u, y-v)-\bar{t})^2}}}
\end{equation}
where 
$\bar{f}_{u,v} = \frac{1}{N_xN_y} \sum_{x=u}^{u+N_x-1} \sum_{y=v}^{v+N_y-1} f(x,y)$
Here, the detected bounding box of the image in the previous frame is used as a template $t$. The next position of the track estimated by the Kalman filter is used as $(u, v)$. Normalized cross correlation is computed between the crop of the incoming frame $f$ centered around the $(u, v)$ and the template $t$.

\noindent \textbf{Histogram of Oriented Gradients (HOG) based Similarity.}
Compared to pixel-wise similarity computation, computing features by aggregating nearby pixels and computing the similarities between the extracted features will allow the algorithm to skip more frames in scenarios where the object/part of the object has moved by a very small number of pixels. Histogram of Oriented Gradients (HOG) \cite{HOG} counts the occurrences of each orientation of gradients in each localized portion of the image. This approach is known to be robust to geometric and photometric transformation but not to object orientation over large spatial regions, which would make it suitable for our usage case.
For images of the same size, HOG computes a feature vector of the same dimension. Computing the cosine similarity between these normalized HOG features can be used as a measure of similarity between two images. This method is found to be effective and still less computationally intensive compared to a DNN based feature extractor.

\subsubsection{State Estimation - Kalman Filter}
Given that the noise is normally distributed, Kalman filter computes the optimal state estimate by combining the previous estimates with new observations. 
Kalman filter is used to estimate the next position of tracklets in the skipped frames in our approach because it is less compute-intensive and the same Kalman filter that is used for the position estimation step of association can be reused with less computing overhead. Kalman filter is also known to work well for linear motion and can make good predictions during periods of occlusions as well and is used for state estimation by most of the object tracking approaches.

\subsection{Algorithm Overview}
The pseudo-code for context-aware skipping using normalized cross correlation as the similarity measure and Kalman filter for state estimation is provided in Algorithm~\ref{alg:skippingAlg1}.

 \begin{algorithm}[H]
 \caption{\label{alg:skippingAlg1} Pseudo-code for Context Aware Skipping (Skipping based on NCC/HOG similarity and next positions  predicted by Kalman filter)}
 \begin{algorithmic}[1]
 \renewcommand{\algorithmicrequire}{\textbf{Input:}}
 \renewcommand{\algorithmicensure}{\textbf{Output:}}
 \REQUIRE Video frames
 \ENSURE  For each frame, list of bounding boxes and their IDs across frames
 \\ \textit{Initialisation}: 
  \STATE Run Detection on Frame 1, initialize tracklets
 \\ \textit{LOOP Process}
  \FOR {each incoming frame $f$}
  
  \STATE templates := image crops containing detections on the previous frame
  \STATE Estimate the next state of all tracklets using the Kalman Filter
  \STATE estimations := image crops with estimated bounding boxes from the current frame
  \STATE similarity measure := average of normalized cross-correlation between all (template, estimation)
  \\(or
  similarity measure := average of cosine similarity between Histogram of Oriented Gradients computed for all (template, estimation))
  \IF {$\textit{similarity measure}$ $NCC \ge 0.75$ (or $HOG \ge 0.85$)}
  \STATE \# Normal process
  \STATE Run MOT detection
  \STATE Estimate the next state of each track
  \STATE Compute association between detected bounding boxes and existing tracks
  \STATE Update Kalman Filter with observations
  \ELSE
  \STATE \# Skip detection on frame $f$
  \STATE Update the tracks' current position with Kalman filter’s predictions
  \STATE Use these predicted values as detections for frame $f$
  \ENDIF
  \STATE Handle track reactivation and lost tracks
  \ENDFOR
 \RETURN Bounding boxes with identities for each frame 
 \end{algorithmic} 
 \end{algorithm}

First, the object detection is performed on the first frame. For the next frame $t$, Lines 3$\sim$7 of the algorithm computes whether to skip detections on the current frame or not. Kalman filter is used to predict the next position of each of the objects (bounding boxes) detected in frame $t+1$ (Line 4). The image crops containing the detected objects in frame $t$ and the image crops based on estimated positions of the objects in frame $t+1$ are taken (Lines 3 \& 5). For each object in frame $t$, the similarity measure (NCC or HOG) between its image crop in frame $t$ and its corresponding estimated image crop in frame $t+1$ (Lines 6) is computed and an average of the measure is computed over all objects. If the similarity computed is less than a pre-defined threshold, the estimated position in frame $t+1$ is considered to be incorrect (could be possible due to occlusion, non-linear motion, etc.), and the normal object detection and association process is performed (Lines 9$\sim$12). If the similarity is greater than a specific threshold, the estimated position at frame $t+1$ is considered to be the correct next position of the object. Detection on that frame will be skipped and the tracklets are updated with the estimated position as the tracking result for frame $t+1$. The same process is repeated for all the frames in the video.

\section{Experimental Analysis}
The proposed context-aware skipping approach can be applied on top of the object trackers that employ the tracking-by-detection paradigm, i.e., separately performing object detection and tracking, such as FairMOT~\cite{FairMOT}.
We implement the proposed context-aware skipping optimizations on top of a state-of-the-art object tracker, FairMOT~\cite{FairMOT}. The experimental evaluations are conducted on MOT-15 and MOT-17 datasets~\cite{MOT15, MOT16} using the Nvidia Tesla K80 GPU on Microsoft Azure.
We consider three baseline methods: (1) no skipping, which is equivalent to FairMOT~\cite{FairMOT}, (2) periodic skipping, where the detections are performed every $\omega$ frames, e.g., $\omega$=2 by default, and the latest detections will be used for the skipped frames, and (3) alternate skipping, which is inspired by HTracker~\cite{HTracker} to detect on every alternate frame and use the Kalman filter's estimate as the detections for the skipped frame.
In order to analyze the impacts of our proposed approach on the object tracking quality, we report the MOTA, MOTP \cite{CLEARmetrics}, ID F1 \cite{IDmetrics}, and ID Switches, which are commonly used in the literature for evaluating multi-object trackers. In addition, we also report the unified metric HOTA \cite{HOTAmetrics} and its sub-metrics DetA, LocA and AssA for the proposed approach and baselines.

\begin{table*}[htbp]
\caption{Comparison of optimization approaches implemented on FairMOT evaluated on MOT-15 dataset (AFD = adaptive forced detections)}
\label{tab:MOT15result}
\begin{center}
\scalebox{0.92}{
\begin{tabular}{|c|c|c|c|c|c|c|c|c|c|}
\hline
\textbf{Approach}&\textbf{MOTA($\uparrow$)}&\textbf{MOTP($\uparrow$)}&\textbf{IDF1($\uparrow$)}&\textbf{IDsw($\downarrow$)}&\textbf{DetA($\uparrow$)}&\textbf{AssA($\uparrow$)}&\textbf{LocA($\uparrow$)}&\textbf{HOTA($\uparrow$)}&\textbf{\#Frames Skipped : \%}\\
\cline{0-9} 
GSDT \cite{GSDT} & 60.7 & N/A & 64.6 & 477 & 51 & 48.5 & 80.4 & 49.5 & 0\%\\
\hline
ReMOT \cite{ReMOT} & 63.6 & N/A & 67 & 445 & 52.3 & 49.2 & 79.4 & 50.6 & 0\%\\
\hline
No skip (Baseline 1) &	66.2 &	79.625	&73.2\% &	146  &	54.088 & 57.598	&82.199	& 55.727&	0\%\\
\hline
\hline
Alt skip (Baseline 2)&	61.6	& 78.046 &	68.1\%&	244&	51.443   & 52.549& 80.845	&	51.9&	2747 : 50\%\\
\hline
Alt skip + estimation &	64.1 & 78.896	& 71.1\% & 156	&	52.755&	56.161&	81.567&	54.361 &	2747 : 50\%\\
\hline
NCC + estimation + AFD (Ours) &	\textbf{65.1}&	\textbf{79.031} &	72.4\% & \textbf{136} &  \textbf{53.353}	&	56.506 & \textbf{81.711} &	54.829 & 2077 : 37.8\%	\\
\hline
HOG + estimation + AFD (Ours) &	\textbf{65.1} &	78.969 &	\textbf{72.8\%} &	144 &	53.349 &	\textbf{56.699} &  81.679	& \textbf{54.917}	&	\textbf{2367 : 43.1}\%\\
\hline
\end{tabular}
} 
\label{tab5}
\end{center}
\end{table*}

\begin{table*}[htbp]
\caption{Comparison of optimization approaches implemented on FairMOT evaluated on MOT-17 dataset (AFD = adaptive forced detections)}
\label{tab:MOT17result}
\begin{center}
\scalebox{0.92}{
\begin{tabular}
{|c|c|c|c|c|c|c|c|c|c|}
\hline
\textbf{Approach}&\textbf{MOTA($\uparrow$)}&\textbf{MOTP($\uparrow$)}&\textbf{IDF1($\uparrow$)}&\textbf{IDsw($\downarrow$)}&\textbf{DetA($\uparrow$)}&\textbf{AssA($\uparrow$)}&\textbf{LocA($\uparrow$)}&\textbf{HOTA($\uparrow$)}&\textbf{\#Frames Skipped : \%}\\
\cline{0-9} 
MotionTrack\cite{MotionTrack} & 81.1 & N/A & 80.1\% & N/A & 65.1 & 65.4 & 83.2 & 65.1 & 0\% \\
\hline
BoT-SORT\cite{BoTSORT} & 80.5 & N/A & 80.2\% & N/A & 64.9 & 65.5 & 83.2 & 65 & 0\% \\
\hline
Deep OC-Track \cite{DeepOCSORT} & 79.4 & N/A & 80.6\% & N/A & 64.1 & 65.9 & 83.4 & 64.9 &0\% \\
\hline
No skip (Baseline 1) &	73.002 &	83.312	&78.2\% &	284& 60.408 &	66.242 &	85.097&	63.178&	0\%\\
\hline
\hline 
HTracker \cite{HTracker} & 66.9 & N/A & 70.4\% & N/A & 55.3 & 55.5 & 81.6 & 55.3 & 50\% \\
\hline
Alt skip (Baseline 2)& 70.8&  82.42& 78.0\% & 432& 58.59& 84.419& 67.136& 62.618 & 2657 : 50\% \\
\hline
Alt skip + estimation &	71.856	& 82.884&	78.5\%&	260&	59.295&	67.277&	84.809&	63.075&	2657 : 50\%\\
\hline
NCC + estimation + AFD (Ours) &	71.1&	82.483&	78.9\% &	192 &	58.496&	67.498&	84.459&	62.759&	3230 : 60.7\%\\
\hline
HOG + estimation + AFD (Ours) &	71.3 &	82.453 &	79.4\% &	196 &	58.627&	 67.919 &	84.478 &	63.027 &	3115 : 58.62\%\\
\hline
\end{tabular}
} 
\label{tab4}
\end{center}
\end{table*}

Table~\ref{tab:MOT15result} and Table~\ref{tab:MOT17result} present the experimental comparison of the representative multiple object tracking methods, baseline methods, and proposed context-aware skipping methods on MOT-15 and MOT-17 datasets respectively.
Compared to Baseline 1 (equivalent to FairMOT) with no skipping, the proposed approach can skip detections on more than half of the frames and only exhibit a small drop ($\approx$ 0.5 – 2\%) on MOTA, MOTP, detection accuracy, and localization accuracy for most of the videos. The proposed approach only experiences a small loss in association accuracy, ID F-1, and reaches a comparable number of ID switches for most of the input videos. 
Compared to alternate frame skipping (Alt skip (Baseline 2)) and estimation baseline (Alt skip + estimation), the proposed approach is able to skip 20\% more frames in MOT-17 and still reach a higher HOTA, ID F-1, fewer ID switches, and a comparable value for MOTA and MOTP.

\begin{figure}[htbp]
\centerline{\includegraphics[width=0.4\textwidth]{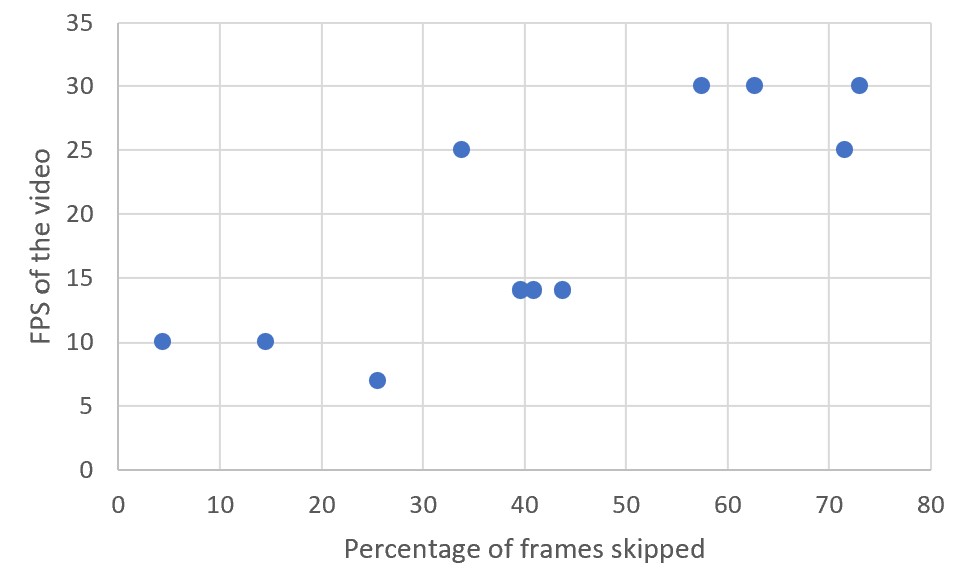}}
\caption{Plot showing the percentage of frames skipped by the NCC + estimation + adaptive forced detections v.s. the FPS of different videos from MOT-15 dataset}
\label{FramesSkippedVsFPS_mot15}
\end{figure}

\begin{figure}[htbp]
\centerline{\includegraphics[width=0.4\textwidth]{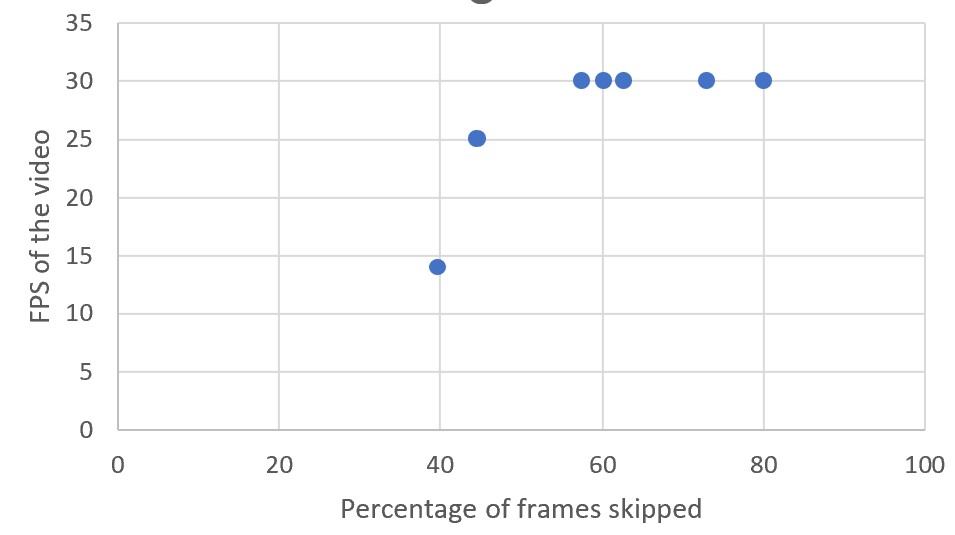}}
\caption{Plot showing the percentage of frames skipped by the NCC + estimation + adaptive forced detections v.s. the FPS of different videos from MOT-17 dataset}
\label{FramesSkippedVsFPS_mot17}
\end{figure}

\begin{figure*}[ht]
  \centering  
  \vspace{-10ex}
  \includegraphics[width=0.97\textwidth, height=0.85\textheight]{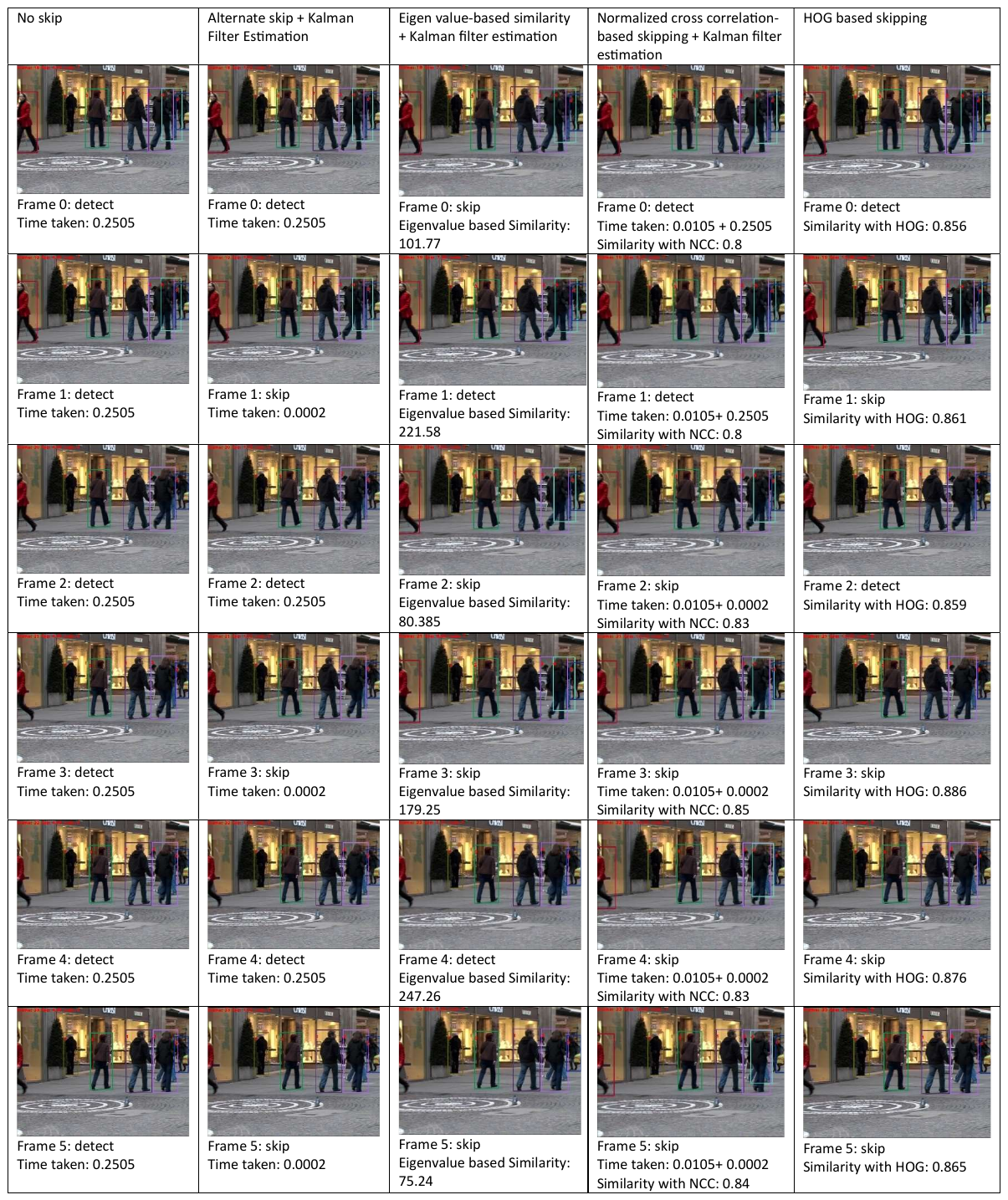}
  \vspace{-18ex}
  \caption{Each column shows the tracking results of six consecutive frames from TUD-Stadtmitte (25 FPS) from MOT-15 dataset when different approaches (periodic, eigen-value, normalized cross correlation, similarity of Histogram of Oriented gradients) are used to determine the frames that can be skipped and when Kalman filter is used to estimate the position of bounding boxes in the skipped frames}
  \label{fig:TUDS_5_sample_frames_dynamic}
\end{figure*}

\begin{figure*}[ht]
  \centering  
  \vspace{-10ex}
  \includegraphics[width=0.97\textwidth, height=0.85 \textheight]{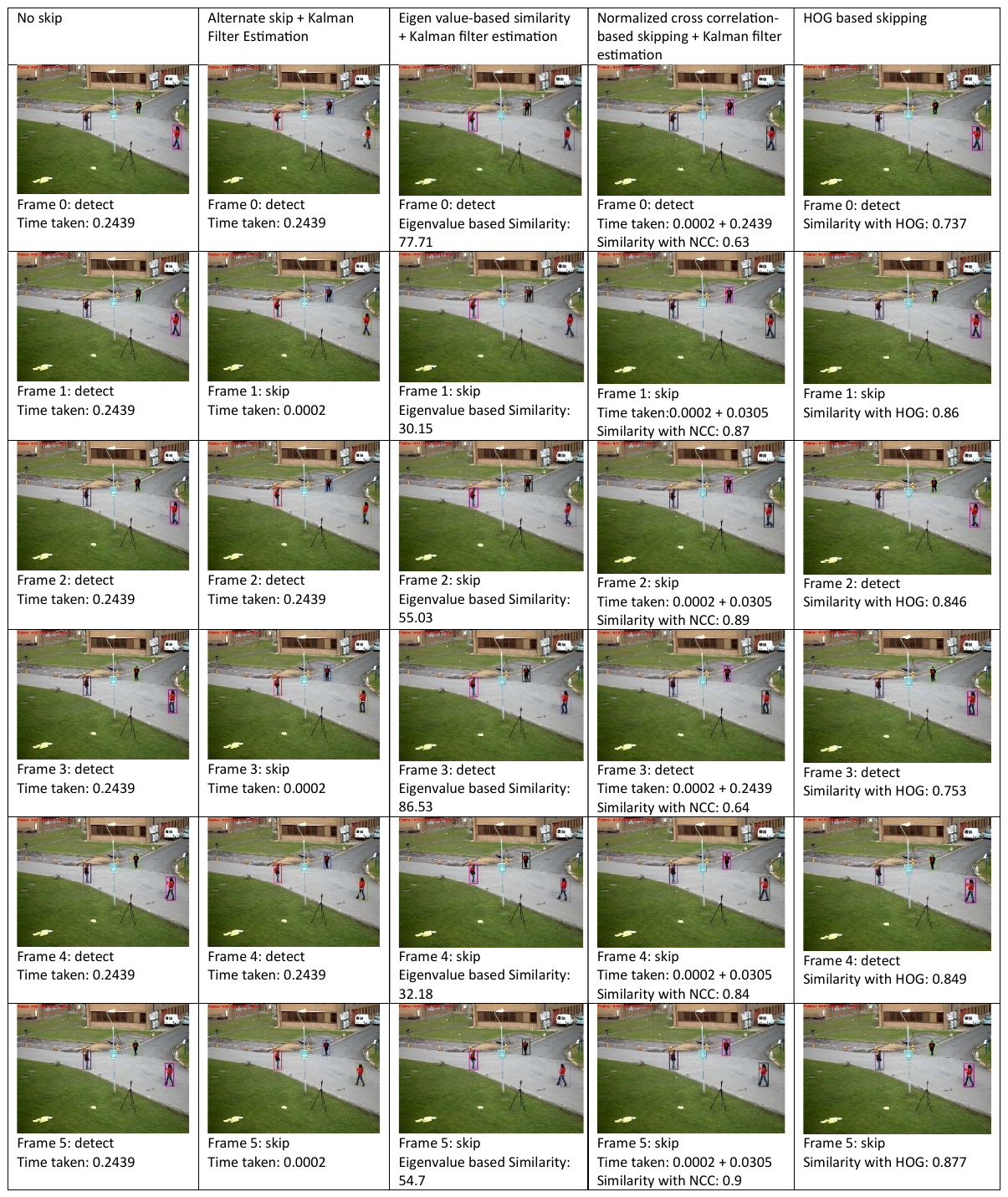}
  \vspace{-18ex}
  \caption{Each column shows the tracking results of six consecutive frames from PETS09-S2L1 (7 FPS) from MOT-15 dataset when different approaches (periodic, eigen-value, normalized cross correlation, similarity of Histogram of Oriented gradients) are used to determine the frames that can be skipped and when Kalman filter is used to estimate the position of bounding boxes in the skipped frames  }
  \label{fig:PETS_5_sample_frames_dynamic}
\end{figure*}

\noindent \textbf{Visualization.} 
Figure~\ref{FramesSkippedVsFPS_mot15} and~\ref{FramesSkippedVsFPS_mot17} show the percentage of frames that the proposed approach was able to skip for videos of different incoming frame rates on MOT-15 and MOT-17 datasets respectively.
We observe that for videos with a higher FPS, more frames are skipped compared to videos with a lower FPS. Moreover, The proposed context-aware skipping approach can decide which frames to skip adaptively based on the motion observed instead of showing a linear relationship between the video FPS and the percentage of frames skipped.
Figure~\ref{fig:TUDS_5_sample_frames_dynamic} and~\ref{fig:PETS_5_sample_frames_dynamic} further visualize multiple object tracking results for TUD-Stadtmitte (25 FPS) and PETS09-S2L1 (7 FPS) videos from the MOT-15 dataset.
We observe that using a Kalman filter to estimate the object positions on these skipped frames improves the localization accuracy as compared to simply reusing the previous detections. In Figure~\ref{fig:TUDS_5_sample_frames_dynamic}, it is noticeable that using the proposed context-aware skipping scheme, over half the number of frames can be skipped in the video when the motion of objects between frames is small and linear.
Overall, our proposed context-aware skipping approach shows better tracking performance compared to skipping periodically with a fixed frequency or using heuristics to determine the skipping frequency.

\begin{table}[htbp]
\caption{Comparison of time taken for detection and skipping + estimation}
\begin{center}

\begin{tabular}{|p{2.5cm}|p{2.2cm}|p{2.2cm}|}
\hline
\textbf{Step} & \textbf{Avg Time taken for MOT-15 (seconds)} & \textbf{Avg Time taken for MOT-17 (seconds) } \\
\hline
Decision to skip (position estimation + NCC) & 0.011596515 & 0.026890381	 \\
\hline
Predict next position (if decision is to skip) & 0.000252304 & 0.000341388	 \\
\hline
Run Detection and update tracks (if decision is not to skip) & 0.246894905 & 0.248551353	 \\
\hline
\end{tabular}
\label{timecomparison}
\end{center}
\end{table}

\noindent \textbf{Runtime Performance.}
We then evaluate the speed-up obtained by the proposed approach using NCC similarity. We consider (1) the average time taken for the computations required to decide whether to skip the frame or not ($t_{decision}$), (2) the average time taken for position estimations of the skipped frame ($t_{estimation}$), and (3) the average time taken to run detection and association ($t_{detection}$) if the frame is not skipped as reported in Table \ref{timecomparison}. For MOT-17 dataset, it is observed that the $t_{decision} = (1/9) * t_{detection}$. When 60\% of the frames are skipped, we have

\begin{equation}
\begin{aligned}
t_{total\_EMO} & = (t_{detection} + t_{decision}) * n_{frames} * 0.4 \\
& \quad + (t_{decision}) * n_{frames} * 0.6 \\
& = 0.51 * t_{detection} * n_{frames}\\
& = 0.51 * t_{total\_noskip}
\end{aligned}
\end{equation}

For MOT-15 dataset, it is observed that the $t_{decision} = (1/20) * t_{detection}$. When 38\% of the frames are skipped, 
\begin{equation}
\begin{aligned}
t_{total\_EMO} &= (t_{detection} + t_{decision}) * n_{frames} * 0.62 \\
& \quad + (t_{decision}) * n_{frames} * 0.38 \\
& = 0.67 * t_{detection} * n_{frames}\\
& = 0.67 * t_{total\_noskip}
\end{aligned}
\end{equation}

Based on the above analysis, the proposed context-aware skipping approach consumes only 50-67\% time to perform multi-object tracking compared to the baseline no skipping approach while the loss in object tracking accuracy (e.g., HOTA) due to skipping is very small ($\sim$0.5\%).

\section{Conclusion}
This paper aims to reduce the computation costs of multi-object trackers by strategically skipping detections without compromising multi-object tracking quality. We make three original contributions.
{\it First,} we present an empirical analysis of the practical issues of the periodical detection skipping scheme with a fixed skipping frequency.
{\it Second,} we propose a context-aware skipping approach to dynamically skip detections on varying numbers of frames depending on the image similarity and object motion predictability.
{\it Third,} we conduct comprehensive experiments on benchmark datasets to evaluate the proposed approach and design alternatives, which demonstrate the effectiveness of our proposed approach in reducing the computational costs and maintaining high quality for multi-object tracking.
The proposed approach is general and works well on top of object trackers that follow the tracking-by-detection paradigm. 

\ifCLASSOPTIONcompsoc
  \section*{Acknowledgments}
\else
  \section*{Acknowledgment}
\fi

This research is partially sponsored by CISCO Edge AI program (2021-2024) and the NSF CISE grants 2038029, 2302720, and 2312758.

\end{document}